\documentclass[acmlarge,screen]{acmart}

\usepackage{subfiles}
\usepackage{dsfont}
\usepackage[normalem]{ulem}
\AtBeginDocument{%
  \providecommand\BibTeX{{%
    \normalfont B\kern-0.5em{\scshape i\kern-0.25em b}\kern-0.8em\TeX}}}

\usepackage{caption}
\usepackage{subcaption}
\begin{document}

\title{Transformers for prompt-level EMA non-response prediction}


\author{Supriya Nagesh}
\affiliation{%
  \institution{Georgia Insistute of Technology, USA}
  }
\email{snagesh7@gatech.edu}

\author{Alexander Moreno}
\affiliation{%
  \institution{Georgia Institute of Technology, USA}
}

\author{Stephanie M. Carpenter}
\affiliation{%
 \institution{University of Michigan, USA}
}

\author{Jamie Yap}
\affiliation{%
 \institution{University of Michigan, USA}
}

\author{Soujanya Chatterjee }
\affiliation{%
  \institution{University of Memphis, USA}
}

\author{Steven LLoyd Lizotte}
\affiliation{\institution{University of Utah, USA}}

\author{Neng Wan}
\affiliation{\institution{University of Utah, USA}}

\author{Santosh Kumar}
\affiliation{\institution{University of Memphis, USA}}

\author{Cho Lam}
\affiliation{\institution{University of Utah, USA}}

\author{David W. Wetter}
\affiliation{\institution{University of Utah, USA}}

\author{Inbal Nahum-Shani}
\affiliation{\institution{University of Michigan, USA}}

\author{James M. Rehg}
\affiliation{\institution{Georgia Institute of Technology, USA}}

\renewcommand{\shortauthors}{Nagesh, et al.}

\begin{abstract}



Ecological Momentary Assessments (EMAs) are an important psychological data source for measuring current cognitive states, affect, behavior, and environmental factors from participants in mobile health (mHealth) studies and treatment programs. \textit{Non-response}, in which participants fail to respond to EMA prompts, is an endemic problem. The ability to accurately predict  non-response could be utilized to improve EMA delivery and develop compliance  interventions. Prior work has explored classical machine learning models for predicting non-response. However, as increasingly large EMA datasets become available, there is the potential to leverage deep learning models that have been effective in other fields. Recently, transformer models have shown state-of-the-art performance in NLP and other domains. \textit{This work is the first to explore the use of transformers for EMA data analysis}. We address three key questions in applying transformers to EMA data: 1. Input representation, 2. encoding temporal information, 3. utility of pre-training on improving downstream prediction task performance.  The transformer model achieves a non-response prediction AUC of 0.77 and is significantly better than classical ML and LSTM-based deep learning models. We will make our a predictive model trained on a corpus of 40K EMA samples freely-available to the research community, in order to facilitate the development of future transformer-based EMA analysis works.
\\
\textbf{Manuscript under review at ACM IMWUT}

\end{abstract}

\maketitle

\section{Introduction}

Mobile health (mHealth) technology is a promising tool for health behavior change and maintenance with a broad array of applications, including smoking cessation \cite{riley2011health}, physical activity \cite{klasnja2019efficacy}, stress management \cite{mayor2015mobile} and medication adherence \cite{park2019mobile}. mHealth 
data sources such as wearable sensors, self-reports, GPS, etc. provide key insights into the contextual and behavioral factors that influence health outcomes, through the ability to collect data from participants in real-time in the field environment. A particularly valuable source of data comes from ecological momentary assessments (EMAs), in which participants answer questions about their mental state, behaviors, and other factors by completing questionnaires, typically multiple times per day. EMA data provides unique insights which are difficult to glean from other sensing modalities, and is widely-used as a result. It can be used to assess the risk of adverse behaviors, trigger interventions, or estimate treatment effects. 

A major challenge in EMA data collection is \emph{participant non-response}, which arises when users fail to complete a survey when prompted. Non-response is problematic for three reasons. First, it reduces the statistical power when testing hypotheses using mHealth data. Second, if the non-response is systematic, then it is likely to be missing not at random (MNAR), a form of bias which is difficult to correct for. Third, missing EMA samples make it more challenging to assess time-varying contextual variables such as emotions, environments, and behaviors. Non-response can have many causes. For example, participants who are driving or in a meeting may not be available to respond to the prompt. Non-response can also arise due to a lack of participant engagement or motivation. A predictive model that could identify moments when non-response is likely would be a useful tool in improving EMA yields. A predictive model could be used by the EMA scheduler to deliver prompts when the likelihood of a response is higher, and it could also be used to trigger additional incentives or compliance interventions designed to improve the EMA response rate.

The goal of this paper is to develop a machine learning model that can predict the future risk of non-response from the history of EMA responses. Prior work on \textit{studying} non-response ~\cite{sokolovsky2014factors,courvoisier2012compliance,turner2017race,ono2019affects,boukhechba2018contextual,mishra2017investigating} has focused on identifying effective predictors using classical machine learning methods. Given the variety of factors that can contribute to nonresponse, a data-driven feature learning approach is attractive. Only a few prior works have used deep learning (DL) for EMA data modeling~\cite{suhara2017deepmood, mikus2018predicting}, in contrast to other mHealth data types such as accelerometry~\cite{chen2015deep,10.1145/2733373.2806333,ronao2016human,yang2015deep}, and no prior works have used DL for non-response prediction. Recently, transformers~\cite{vaswani2017attention} have emerged as a powerful new class of tools for modeling sequential observations. Following their initial success in NLP~\cite{devlin2018bert, radford2018improving}, transformers have proven effective in computer vision\cite{girdhar2019video} \cite{ranftl2021vision} \cite{khan2021transformers} and speech recognition~\cite{Paraskevopoulos2020}, among other domains. Sequences of EMA observations differ significantly from NLP and time series data in that the arrival times are \emph{irregularly-sampled} and important to model (e.g. EMA responses which are closer together in time are more likely to be correlated). \emph{We are the first to explore the use of transformers for EMA data analysis in general, and non-response prediction in particular.}

We address three issues in applying transformers to EMA data modeling. The first is the choice of input representation. While many applications require domain-specific input embeddings (e.g. word embeddings in the case of NLP), we find that the fixed length EMA response vector itself is an effective input representation. The second issue has to do with the method for encoding observation times. Positional encoding introduced in~\cite{vaswani2017attention} adds a vector to each input embedding which provides a global encoding of the position of each word. Similarly, we find that explicitly encoding the EMA times along with the responses improves performance. However, we find that concatenating the positional encoding is more effective than adding it. The third issue is the utility of pre-training in improving prediction performance. The BERT architecture for NLP tasks~\cite{devlin2018bert} demonstrated the effectiveness of pre-training a transformer-based model on a large unlabeled corpus prior to fine-tuning it with labels on a smaller task-specific training dataset. We designed a self-supervised pre-training task based on EMA imputation and evaluated its effectiveness for non-response prediction. We found that pre-training produced a small performance benefit which was not statistically-significant. We hypothesize that this approach may be more effective in the future as larger EMA datasets become available. We present visualizations of the learned transformer representation that suggest that it encodes structure in the EMA response data which is meaningful for non-response prediction.

In summary, this paper makes the following contributions: 

\begin{itemize}
    \item This is the first work to explore the utility of transformer models for sequences of EMA response data. We present a transformer architecture for predicting non-response to EMA prompts using the history of EMA responses. The transformer model achieves an AUC of 0.77 for predicting future non-response and is significantly more accurate than both classical ML models and DL models based on the LSTM architecture.
    \item We present the design decisions that yield effective transformers for EMA sequence analysis, investigating input and positional encoding methods and identifying the most effective strategies. We present visualizations that illustrate the ability of the transformer to learn meaningful representations for non-response prediction.
    \item We design a self-supervised pre-training task to learn the structure within EMA sequences. We evaluate the utility of pre-training and report a modest performance gain which is not statistically-significant.
\end{itemize}

\section{related work}

There are three bodies of prior work which are most closely-related to this paper: 1) Analysis and prediction of EMA non-response, as well as the related topics of interruptability and availability;  
2) Use of deep learning models to analyze EMA data; and 
3) Transformer models for electronic health record (EHR) data, which shares with our task the need to model irregularly-sampled data. We discuss each of these topics in detail.

\subsection{Analyzing and predicting EMA non-response}

A significant body of prior work analyzes the factors that are related to non-response to EMA prompts. \cite{sokolovsky2014factors,courvoisier2012compliance} identify the factors that have a significant effect on non-response (which they term compliance, adherence, engagement). \cite{shiyko2017feasibility, turner2017race} study EMA response rates and determine the feasibility of using EMA as a research tool based on the response rates. \cite{shiyko2017feasibility} further underscores the importance of differentiating between human factors and factors related to technology in non-response while reporting response rates (which they refer to as adherence level). Two recent review papers on EMA non-response, \cite{jones2019compliance} and \cite{ono2019affects}, provide additional evidence for the importance of the problem. \cite{ono2019affects} reviews studies involving patients with chronic pain, while \cite{jones2019compliance} reviews studies related to substance abuse. In contrast to the current study, these prior works do not address the development of a \textit{predictive model} for non-response to EMA.

Two recent works \cite{mishra2017investigating,boukhechba2018contextual} have focused on \emph{predicting} participant non-response. Both works use contextual factors (such as location, activity, etc.) in a predictive model. One common factor among all these prior works is their use of classical machine learning models for analyzing and predicting EMA non-response. We share with these prior works an investigation into the predictive power of various contextual factors and mental states (e.g. emotions). At the same time, our work is uniquely-distinguished by its focus on developing transformer models for non-response prediction in order to exploit the benefits of feature learning in modeling complex sequential data.

The tasks of assessing interruptibility and availability in mHealth are related to our problem of non-response prediction. A representative example of availability modeling is Sarkar et. al.~\cite{sarker2014assessing}, which developed a classifier that combined mobile sensor data with past EMA responses to classify whether or not a participant is available at the current moment in time. The topic of interruptibility has been widely-explored in the context of intelligent notification systems \cite{aminikhanghahi2019context,mehrotra2017intelligent,ho2017emu,ho2020quick}. The goal of these works is to design a system that delivers notifications at opportune moments based on contextual factors. The focus of \cite{mehrotra2017intelligent} is the optimization of the user experience. \cite{ho2020quick} presents a reinforcement learning based method for scheduling notifications. This is similar to the study of receptivity to mHealth interventions in \cite{kunzler2019exploring,choi2019multi}, where the goal is to determine opportune moments using contextual factors (such as activity, location, phone battery, etc.). The topics of availability, receptivity, and interruptibility prediction are critically important for avoiding unnecessary participant burden and considering external contextual factors in determining availability. In contrast, our focus is on developing a predictive model for non-response based on feature learning derived from factors such as participant mental states and emotions, and their history of EMA responses. 

An additional related topic is participant disengagement, which manifests as a steady decline over time in the participation of a user in a study or treatment program~\cite{clawson2015no,lazar2015we}, often resulting in loss to follow-up~\cite{druce2019maximizing}. The focus of these works is on longitudinal analyses and long-term study outcomes. In contrast, our focus is on quantifying the short-term risk for non-response at the EMA prompt level. Such a capability could inform the design of interventions to maximize the utility of EMA as a measurement tool, which is distinct from the important task of improving long-term participant engagement.


A final related topic is in the domain of ePROs (electronic patient-reported outcomes). ePROs are patient-provided information about symptoms, side effects, drug timing and other questions during a clinical trial \cite{coons2015capturing}. ePROs generally lack the momentary, frequent sampling found in our EMA dataset. The extension of our work to developing transformer models for sequences of ePRO data is an interesting avenue for future work.

\subsection{Deep models for EMA data}

There are two prior works that develop deep models for prediction tasks using EMA data~\cite{suhara2017deepmood, mikus2018predicting}. In \cite{suhara2017deepmood}, the authors propose a recurrent neural network {(RNN)} for forecasting depressed mood using the history of EMA data. In \cite{mikus2018predicting}, the focus is on predicting short term mood developments from EMA data using an RNN. In addition, there are numerous works that analyze EMA data using classical statistical and machine learning tools, such as logistic regression and SVMs~\cite{spanakis2016enhancing,saha2017inferring,kim2019depression,van2020predicting,king2018predicting}. The current article differs from these prior works in two ways. First, we address the problem of predicting whether the next prompt will result in an EMA response, which is distinct from the task of predicting the responses themselves, as in the case of predicting self-reported mood. Second, we develop a \textit{transformer model for EMA} sequences and analyze its utility for predicting EMA non-response. Transformer models have been shown to deliver state of the art results in fields such as NLP \cite{vaswani2017attention}\cite{devlin2018bert} and computer vision. We extend this class of models to the EMA setting.


We note that there has been significant work on using DL models to analyze  clinical data such as Electronic Health Records (EHR), a domain with some similarity to EMA analysis. While EHR data is diverse, it includes categorical variables that capture clinical states, which is analogous to EMA response data. Two representative works that use classical sequential DL models for EHR analysis are~\cite{choi2016retain,kaji2019attention}. Both works use an attention layer with a recurrent temporal model (an RNN) for EHR sequence analysis. In contrast, our focus is on the exploration of transformer-style models for irregularly-sampled EMA data.

\subsection{Transformers for Electronic Health Records Data}

Based on the success of transformer models on NLP tasks \cite{vaswani2017attention}\cite{devlin2018bert}\cite{tay2020efficient}, recent works have explored their application to a broad range of other domains, including the analysis of EHR data. EHR analysis includes several prediction tasks, such as length of stay, mortality, and sepsis onset, which share our focus on predictive modeling from irregularly-sampled data. One representative work is \cite{song2018attend}, which applies transformer models to irregularly sampled clinical data. 
In \cite{li2020behrt}, a BERT-style model is developed using a pre-training task that is appropriate for irregularly sampled diagnosis codes.



There are several significant differences between EHR and EMA analysis. First, EHR datasets consist primarily of categorical observations (e.g. diagnostic codes) and real-valued biomarker measurements, while EMA data consists primarily of ordinal vectors. Second, in EHR datasets only a subset of possible observations are available at any point in time, whereas for EMA it tends to be all or nothing (participants either respond to the prompt and answer all of the items or fail to respond at all). Third, EHR data contains many more variables and data item types in comparison to EMA. Given these differences, it is unlikely that findings from modeling EHR data will transfer in any significant way to EMA data analysis.


\section{study protocol and dataset}

The dataset {comes from} a study that examines the influence of intrapersonal and contextual factors on smoking lapse among African American smokers. Data was collected from multiple modalities including EMA prompts, on body sensors, and location from GPS.


{The study participants carried a smartphone provided to them with the study software installed.} {The mobile app} delivered EMA prompts and collected real time continuous data in the participant’s natural environment from multiple sensors. Data was processed in real time on the smartphone and machine learning algorithms were used to extract biomarkers corresponding to specific behavioral and physiological indicators of smoking and stress. In this analysis, we focus on the EMA data, as this provides a rich set of items that capture aspects of contextual and mental state, and is also the most widely-collected datatype in health applications. 
 
We now describe the EMA collection process. In order to begin and end triggering EMAs for the day, participants had to press a button indicating start and end of day. Participants were prompted {by the phone app }to complete three types of Ecological Momentary Assessments (EMAs) on their smartphones during the study - random EMA, stress triggered EMA, smoking triggered EMA. On each day, a participant was prompted with {an average of four} random EMAs. After the day start button was pressed, the day was divided into 4 equal blocks of time. In each block, the phone app checked for the `participant availability,' determined by the battery level (being above 10\%), whether the participant was driving, and if the participant had enabled a `do not disturb' option. The `do not disturb' mode could be used by participants to stop receiving any EMA prompts when they were unavailable. The data collected from the sensors was used to determine smoking events and events of stress. In case of these events, the phone app checked for the same conditions for firing an EMA and triggered a smoking EMA or stress EMA. \textit{In our work, we are interested in predicting non-response to the random EMAs.} 

\begin{figure}
    \centering
    \includegraphics[scale=0.3]{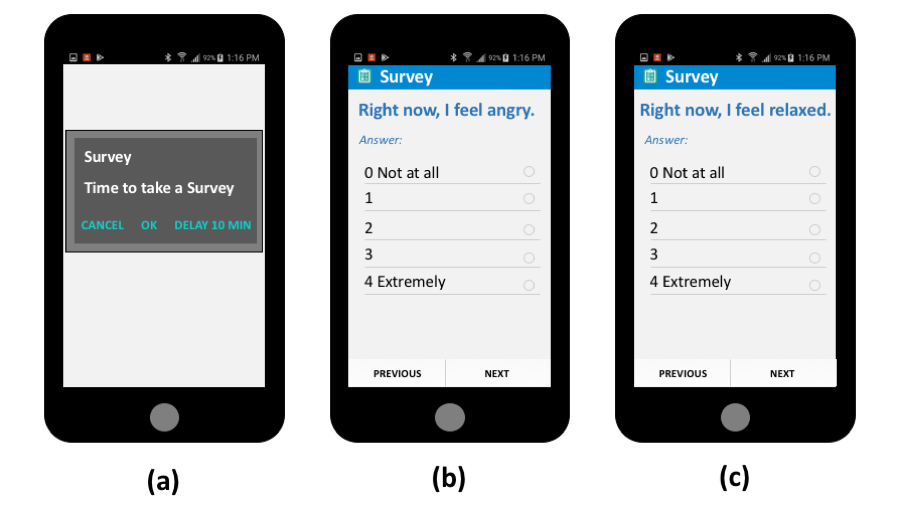}
    \caption{(a) EMA notification on the study phone (b) Survey question: angry (c) Survey question: relaxed.}
    \label{fig:ui}
\end{figure}

{Figure \ref{fig:ui} shows the interface for an EMA notification and the UI while responding to some example survey questions. Once a notification was triggered, the participant could either: 1. Accept the notification and begin answering the survey by clicking `OK', 2. Dismiss the notification by clicking `Cancel', 3. Snooze the notification and receive it again after 10 minutes.}

The dataset consists a total of 255 participants, after excluding participants who dropped out of the study. The participants range between age 20 to 82 (mean 51 $\pm$ 12 years) and we have a roughly balanced split between the male and female subjects. The data collection process spanned two contiguous weeks (4 days pre-smoking-cessation through 10 days post-smoking-cessation). Over the course of the study a total of 9043 random EMAs were triggered and 5636 of them were completed (average compliance rate of 62.8\%).



\section{methodology}

\subsection{Non-response prediction problem framing}
For our analysis, we are interested in predicting response to random EMAs. From here on, we use the term EMA to refer to random EMA. 

\begin{figure}
    \centering
    \includegraphics[scale=0.45]{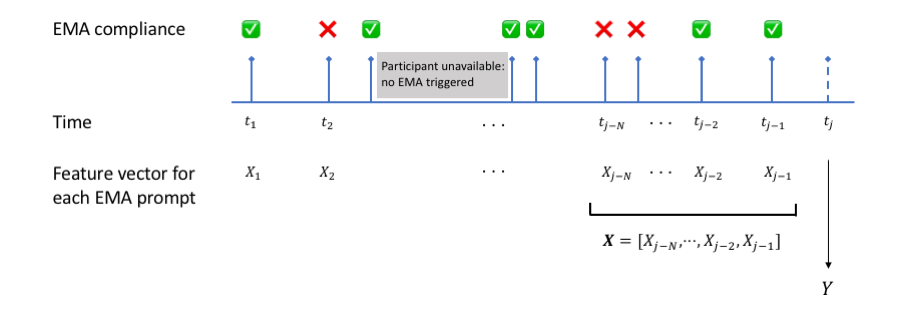}
    \caption{Sliding window approach used to create our feature vector sequence and classification label. N is the sequence length and feature vectors from N consecutive EMAs are concatenated and the prediction label $Y$ is the compliance to the $(N+1)^{\textrm{th}}$ EMA.}
    \label{fig:sliding_window}
\end{figure}

Consider a set of $n$ participants indexed as $i = 1, 2, \cdot \cdot \cdot ,n$. Each participant then has EMAs (observations) indexed by $j = 1, 2, \cdot \cdot \cdot, n_i$, where $n_i$ is the number of observations (EMAs) for participant $i$. We design a model to use a sequence of N EMAs as input and predict if the (N+1)$^{th}$ EMA is completed. See Figure~\ref{fig:sliding_window} for an illustration of the setup. We frame this as a binary classification problem where our label is 
\begin{equation*}   
Y_{j}  = \begin{cases} 
1 & \text{if $j^{th}$ EMA is completed}\\
0 & \text{if $j^{th}$ EMA is missed}\\ 
\end{cases}
\end{equation*}

The feature vector derived from the $j^{th}$ EMA is denoted as $X_j$. Section \ref{sec:feature_const} describes the process to create $X_j$ from the EMA data. Here, we assume that $X_j$ is a feature vector consisting of $K$ features. 


We developed models under two scenarios. In the first, we used a sequence length of one, meaning that for each EMA we predict the compliance for the next EMA prompt. In the second scenario, we used a sliding window (of length $N$) approach as shown in Figure \ref{fig:sliding_window} to compute the feature sequence and the corresponding binary label for classification. In this case, we use a transformer architecture to perform feature learning, and predict next EMA compliance from a sequence of feature vectors. For our experiments, we use a sequence length N = 5, 10, 15, and 25.

\subsection{Feature construction}
\label{sec:feature_const}

We now describe our approach to feature engineering for the task of compliance prediction. We use a set of raw features derived directly from the EMA response along with meta-data logged as part of the study. These are listed in Table \ref{tab:ema_feat} under the column `Features'. The raw features are obtained from the following sources:
\begin{enumerate}
    \item Positive and negative affect, smoking urge: Self-reported  
    \item Time, compliance: EMA logs
    
\end{enumerate}

Additionally, we construct summary features using the history of the raw features. These are listed in Table \ref{tab:ema_feat} under the column `Summary features'. There are two types of summary features constructed, they are designed to: 1) Capture the completion history summary (long term and short term); 2) Capture the variance in the positive and negative affect, and completion pattern.  

The long-term completion rate feature is computed to capture the trait of the participant. Trait refers to the baseline trend of compliance for that participant. Here, we compute the ratio of total EMAs completed over the total number of EMAs triggered until the current EMA. The short-term completion rate feature, which captures the state of the participant is computed by the ratio of the number of EMAs completed on the previous day to the total number of EMAs triggered on the previous day. The state refers to temporally local changes in the compliance, and this computed feature attempts to capture the state for each participant.  

\begin{equation*}
    \text{Long-term completion rate (CR)} = \begin{cases}
    \frac{\sum_{i=1}^{j}{Y_i}}{j} &\text{if $j \neq 0$}\\
     0 &\text{if $j=0$}
    \end{cases}
\end{equation*}

\begin{equation*}
    \text{Short-term completion rate (CR)} = \begin{cases}
    \frac{\#\text{EMA completed on day $(d-1)$}}{n_{d-1}} &\text{if $n_{d-1} \neq 0$} \\
     0 &\text{if $n_{d-1} = 0$}
    \end{cases}
\end{equation*}
 where $d$ is the day the current EMA is triggered, $n_{d-1}$ is the total number of EMAs triggered on day $d-1$. 

The variance feature is computed for the positive and negative affect and smoking urge. The variance feature for each covariate for the $j^{th}$ EMA is computed as the variance of the covariate until the $j^{th}$ EMA. For example, the variance of the positive affect `Happy' computed for the $j^{th}$ EMA is the variance in response to the question `Happy' for EMA 1 to EMA $j$.

\begin{table}[!htbp]
    \centering
    \begin{tabular}{|c|c|c|c|}
    \hline
          Type & Features & Value & Summary features\\
          \hline
          \hline
          {}&Enthusiastic&{}&\\
          {Positive affect}&Happy &{Likert scale (1-5)}&Variance of each item\\
          {}&Relaxed&{}&\\
          \hline
          {}&Bored & {}&\\
          {}&Sad & {}&{}\\
          {Negative affect}& Angry & {Likert scale (1-5)}&Variance of each item\\
          {}&Restless & {}&\\
          {}&Urge & {}&\\
          \hline
          Compliance & Current EMA status & Binary & Long term CR \\
           & & & Short term CR \\
          \hline
        
    \end{tabular}
    \caption{List of all features used. The raw features derived from each EMA prompt are listed in the column `Features'. We compute additional summary features from the history of the raw features. These are listed in the column `Summary features'.}
    \label{tab:ema_feat}
\end{table}

\subsection{EMA Transformer model}

Transformer models are very popular sequence models for NLP tasks \cite{tay2020efficient}\cite{vaswani2017attention}\cite{devlin2018bert}. In our work, we are modeling a sequence of EMA responses using this class of models. We would like to answer the following questions in the context of EMA data:

\begin{itemize}
    \item How are the EMA responses represented in the transformer?
    \item How is the positional information handled? EMAs are irregularly sampled through the day. How can this information be captured in the model? 
    \item What kind of pre-training tasks can be designed to learn the structure of EMA sequences?  
\end{itemize}

We describe some background about transformers in NLP below. We then describe at our attempt at answering the questions in the context of EMA sequences. 

\subsubsection{Background}
\label{sec:background}
The transformer is a sequence modeling architecture based entirely on attention proposed in \cite{vaswani2017attention}. 
A self-attention mechanism is a mapping between pairs of words in a sentence/input points in a sequence to the output. The scaled dot product attention mechanism introduced in \cite{vaswani2017attention} is computed as
\begin{equation*}
    \text{Attention}(Q,K,V) = \text{softmax}\left(\frac{QK^\top}{\sqrt{d_k}}\right)V
\end{equation*}

where $Q,K,V$ are the query, key, and value matrices computed as a projection of the input sequence $X$ into query, key and value spaces. $Q = X W^Q, K = X W^K, V = X W^V$.  The matrix multiplication $QK^\top$ computes pairwise inner products between every query and key vector pair. The value vector is weighted by this attention matrix. 

Multihead attention performs the attention mechanism described above in $h$ different feature spaces, where $h$ is the number of heads. The attention is computed on the key, query, value matrices projected with $h$ different learned projections and concatenated together. 

\begin{equation*}
    \text{Multihead}(Q,K,V) = \text{Concat}(head_1,head_2, \cdot\cdot\cdot head_h) W^o
\end{equation*}
where $head_i = \text{Attention}(QW_i^Q, KW_i^K, VW_i^V)$ and $W^o$ projects the concatenated output back to the original size. 

Two initial operations are performed on a sentence prior to computing attention:

\begin{enumerate}
    \item Input embedding: learned embeddings are used to convert words to vectors of dimension $d_{model}$.
    \item Positional encoding: since the transformer model does not contain any form of recurrence, information about the position of different words is added to the input representation. Sine and cosine embeddings are computed as shown below and added to the input representation. Here $pos$ corresponds to the position of a word in a sentence.
    
    \begin{align*}
            PE_{(pos,2i)} &= sin(pos/1000^{2i/d_{model}}) \\
            PE_{(pos,2i+1)} &= cos(pos/1000^{2i/d_{model}})
    \end{align*}
   
\end{enumerate}

\subsubsection{EMA setting}

One way to think about the scaled dot product attention, discussed in \cite{tsai2019transformer} is noting we can think of $\exp(QK^T)$ the numerator of the softmax as evaluating kernels between a set of query and key points. This then gives us the interpretation that self-attention can be decomposed into several steps: 1) using the keys to construct a kernel average smoother of the value function, where the domain of this function is a vector summarizing \textit{both} observations and times (via the positional embeddings) 2) evaluating the kernel smoother at the query points in order to obtain a fixed length vector 3) using that fixed length vector as input to a neural network. For EMAs, one can think of this as constructing a kernel regression function to describe the EMA observation trajectory via the keys and values and evaluating this function using the observed EMAs and their time points via the queries. This then gives a summary of the EMA history in a fixed length vector for input into a neural network.

There are two main differences between a sequence of words and a sequence of EMA responses: 1. EMA responses are ordinal and responses are already in a vector form, 2. Continuous time associated with EMA responses: e.g., a sequence of 4 EMAs could have been completed at 10 AM, 11.30 AM, 3 PM, 4 PM respectively. We account for these two differences architecturally in this manner:

\begin{enumerate}
    \item Input embedding: The feature vector computed for each EMA is used directly as the input embedding. For a sequence of N EMAs, we represent the input embeddings as $X_1, X_2, \cdot \cdot \cdot X_N$.   
    
    \item Positional/time encoding: A sequence of EMAs has a sequence of discrete positions and continuous times associated with it. For example, for a sequence of N EMAs with input embeddings $X_1, X_2, \cdot \cdot \cdot X_N$, the corresponding discrete positions are $1, 2, \cdot \cdot \cdot N$ and the continuous time associated is $t_1, t_2, \cdot \cdot \cdot t_N$. We encode this information into the input to the self-attention mechanism. We compare two different ways of encoding the position/time representation:
    \begin{enumerate}
        \item Addition: The embeddings from the position/time values are computed using the sine and cosine functions described in Section \ref{sec:background}. Given a sequence of EMA input embeddings $X$, position $pos$, time $t$, the input to multihead attention when encoding the discrete position and continuous time values respectively are:
        
        \begin{align*}
            \text{Input to multihead attention}^{\text{pos}} &= X + PE(pos) \\
            \text{Input to multihead attention}^{\text{time}} &= X + PE(t)
        \end{align*}
        
        \item Concatenation: The embeddings from the position/time values are computed and \textit{concatenated} with the input embeddings. Given a sequence of EMA input embeddings $X$, position $pos$, time $t$, the input to multihead attention when encoding the discrete position and continuous time values respectively are:
        
        \begin{align*}
            \text{Input to multihead attention}^{\text{pos}} &= \begin{bmatrix}
              X\\
              PE(pos)
            \end{bmatrix}\\
            \text{Input to multihead attention}^{\text{time}} &= 
            \begin{bmatrix}
              X\\
              PE(t)
            \end{bmatrix}
        \end{align*}
        
        where 
        $\begin{bmatrix}
              X\\
              PE(t)
            \end{bmatrix} = \begin{bmatrix}
           X_{1} \\
           PE(t_1)
         \end{bmatrix}, 
         \begin{bmatrix}
           X_{2} \\
           PE(t_2)
         \end{bmatrix}, \cdot \cdot \cdot, 
         \begin{bmatrix}
           X_{N} \\
           PE(t_N)
         \end{bmatrix}$
    \end{enumerate}
    
\end{enumerate}




\subsection{Self-supervised pre-training for transformer}

The goal of pre-training a transformer model in a self-supervised manner is so that it can learn the structure in the data, which can be useful for other downstream tasks. In EMA, we might be interested in some particular prediction problem like predicting the probability of a person drinking alcohol. Since we are limited by the amount of labeled data, pre-training aims to leverage self-supervised learning from a lare corpus of EMA data. If we have a large EMA corpus, we can train a model that can learn the structure between different EMA items and the temporal structure in the data. The model can then be fine-tuned for the prediction task that we are interested in. In this paper, we are evaluating the utility of self-supervised pre-training of transformers with EMA data. Based on the findings from BERT \cite{devlin2018bert}, we envision that self-supervised pre-training might be an attractive strategy.

\label{sec:pt}
\subsubsection{Background}
\cite{devlin2018bert} introduced BERT, which is designed to pre-train bidirectional representations from a large corpus of text in a self-supervised manner. This is done by first pre-training BERT (a bidirectional transformer model) on two tasks: 1. Masked language modeling (MLM), 2. Next sentence prediction (NSP). In the MLM task, some words in the input sentence are replaced by a MASK token. \textit{The model is trained to impute these words correctly}. In the NSP task, a pair of sentences is provided as the input and the model is trained to recognize if the second sentence is a valid `next sentence'. The exact description of the pre-training can be found in \cite{devlin2018bert}.   

The idea behind pre-training BERT in this manner is to learn the structure in language: structure of words within a sentence (MLM) and structure at the sentence level (NSP) without having any specialized labels. Once it has learned the structure, a pre-trained BERT can be used with an additional linear layer for other downstream tasks. 

\subsubsection{Pre-training: EMA transformer}
We design a masked EMA imputation task that similar to the MLM task. The features described in Section \ref{sec:feature_const} are constructed for each EMA prompt. We then obtain fixed length sequences of EMA features using a sliding window. Let the sequence of EMA features be $X_1, X_2, \cdot \cdot \cdot X_N$ where $N$ is the sequence length and $X_i$ is the feature vector computed for the $i^{th}$ EMA in the sequence. $X_i$ consists of features corresponding to the positive and negative affect (emotion items) and compliance history. 

The goal of the masked EMA imputation task is to mask out responses to some emotion items in the sequence, and learn to reconstruct the response to these items. This will help the model learn the structure between the different emotion items and their temporal pattern. The number of emotion items masked is pre-determined by us and the positions where the response is masked is chosen at random. We mask out the emotion item(s) in 15\% of the sequence, determined randomly. However, the masked imputation task is for pre-training purposes only. The input sequence does not contain mask tokens during a downstream fine-tuning task. To account for this, after a particular position is chosen for masking, we replace the input with the mask token 80\% of the time. The  input value is retained as is 10\% of the time and changed to a random value 10\% of the time. This is similar to the masking approach in MLM in \cite{devlin2018bert}.

\begin{figure}[ht]
    \centering
    \includegraphics[scale=0.5]{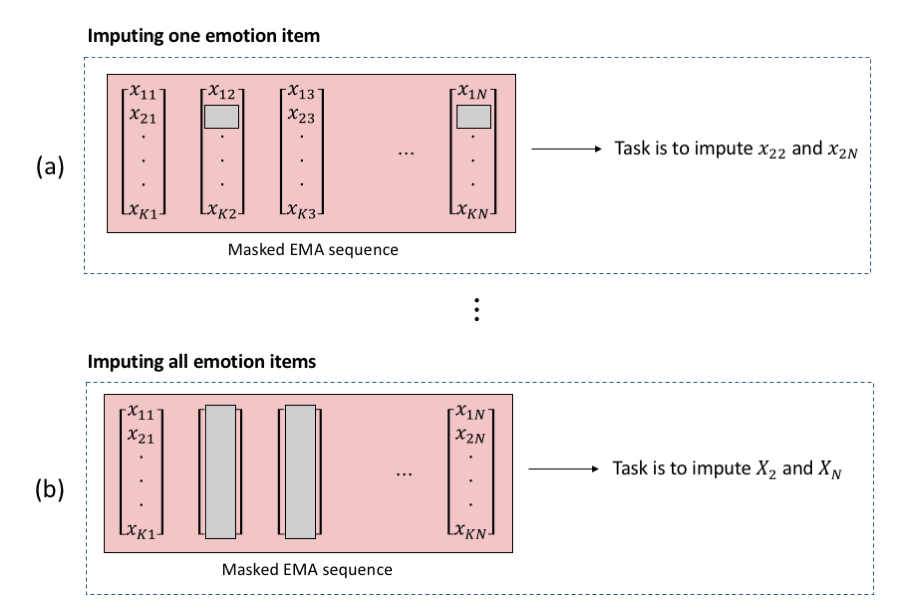}
    \caption{Self-supervised EMA masked imputation tasks. Here we assume that there are $K$ emotion items in each EMA. The input sequence here is depicting \textit{only} the responses to emotion items. We do not mask out other features in the input sequence. (a) The first task shown is to impute one emotion item at a time in 15\% of the sequence positions that are randomly masked. For example, the value of the emotion `Happy' can be masked off in some positions of the sequence and the task is to impute this value correctly. Note that we explore imputing each emotion item one at a time as a pre-training task and evaluate the downstream non-response prediction performance. \textit{There are intermediate tasks possible such as masking out 3 emotion items, 4 emotion items, etc.} (b) The second task is to impute all emotion items in 15\% of the sequence.  }
    \label{fig:pretraining}
\end{figure}

For example, consider EMA sequences of length 10. Suppose we choose to mask out the emotion item `Happy', we randomly choose 2 positions in the sequence at random. For each position with probability 80\%, we replace the EMA response to the question `Happy' in the input sequence with the mask token. The EMA transformer model is then trained to impute the actual value of the response to the question `Happy' in these places. A model pre-trained in this manner will learn the structure in the EMA data and across the emotion items. There are several pre-training tasks possible based on the choice of the emotion items we mask out. Figure \ref{fig:pretraining} shows the two extreme ends of tasks possible. In the first case (a), one emotion item is masked out. In the second case (b), all the emotion items are masked out. The model then has to reconstruct responses to all questions in the masked positions. There are also intermediate tasks possible where we mask out some of the emotion items.  Note that the values masked are always the response to emotion items. We do not mask out the compliance history results for the masked imputation task. 

\begin{figure}[h]
    \centering
    \includegraphics[scale=0.5]{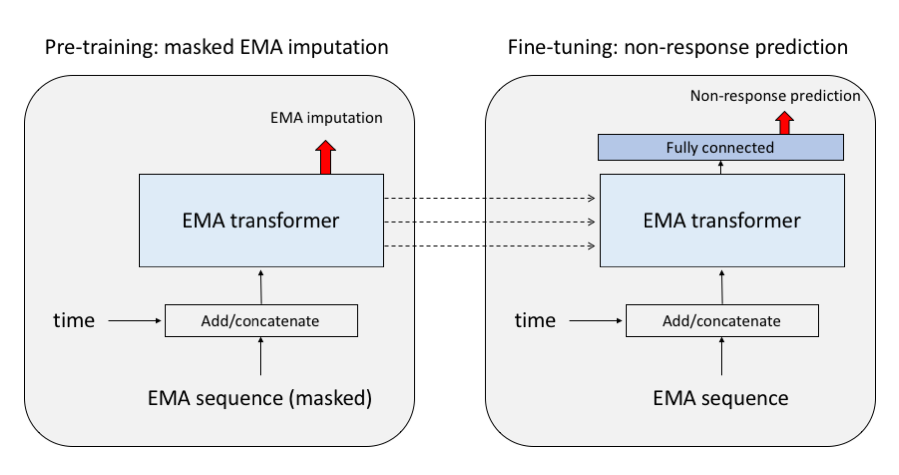}
    \caption{Pre-training the EMA transformer model with an EMA imputation task. The pre-trained model is then fine-tuned for the non-response prediction task. This pre-training strategy is similar to BERT where the model is pre-trained in a self-supervised manner on a large text corpus and the model is fine-tuned for downstream tasks.}
    \label{fig:bert_PT}
\end{figure}

Once we pre-train an EMA transformer model, we add a linear layer to it and fine-tune it for the non-response prediction task as shown in Figure \ref{fig:bert_PT}.

\subsection{Baseline models for comparison}
We consider two non temporal models - Logistic regression, Support Vector Machine (SVM) and compare their performance to ours. We also compare the performance of our model to a vanilla LSTM and an attention LSTM \cite{kaji2019attention} architecture. 

    
    
    
    
    

\subsection{Model development}
{Our model was implemented in PyTorch. The input data was converted to a 3D format with the dimensions as number of subjects, number of time steps, feature input size. We validate the model and perform grid search for hyperparameter tuning using a separate validation data set (containing data from 10\% of the subjects). The model architecture is the standard transformer \cite{vaswani2017attention} encoder architecture. Our architecture consists of 6 encoder layers, 8 attention heads per encoder layer. The dimension of the key, query and value vectors is 64. The scores reported in the results section include 5 fold cross validation results for prediction.}

\section{Experimental results}
\label{sec:results}

In our analysis, we report performance by splitting the data across subjects. We perform 5 fold cross-subject validation, by training the model on data from a set of subjects and testing on data from the held out subjects.

\subsection{Predicting non-response to the next EMA using the current EMA response}

We first evaluate the performance of predicting next EMA compliance when the sequence length N = 1 using non temporal models(using features from one EMA to predict compliance to the next). We perform prediction using two sets of features: 1) Raw features only, 2) raw features and summary features which are described in Table \ref{tab:ema_feat}. The area under the ROC curve (AUC) score for this analysis is presented in Table \ref{tab:baseline_removefeat_nextema}. We see that including the computed summary features result in improved performance. The summary features capture the summary of the \textit{previous} EMA responses. \textit{In the next subsection, we report the performance modeling a sequence of EMA responses to better capture the history. }

    \begin{table}[h]
        \centering
        \begin{tabular}{|c|c|c|}
        \hline
            Model & Raw features only & Raw features \& summary features \\
            \hline
            Logistic regression & $0.63 \pm 0.02$ & $0.71 \pm 0.02$ \\
             \hline
            SVM (RBF kernel) & $0.64 \pm 0.02$ & $0.71 \pm 0.02$ \\
             \hline
        \end{tabular}
        \caption{Average 5 fold cross validation AUC for predicting next EMA compliance using features from the current EMA (N = 1). We use two sets of features: 1) raw features only, 2) raw features and summary features. These features are listed in Table \ref{tab:ema_feat}. }
        \label{tab:baseline_removefeat_nextema}
    \end{table}

\subsection{Predicting non-response to the next EMA using a sequence of EMA responses}

We present results for predicting non-response to the next EMA prompt using a sequence of EMA responses. The features described in Table \ref{tab:ema_feat} corresponding to each EMA is computed. We then use a sliding window to obtain EMA sequences of a fixed length (N)  and the non-response label for the next EMA prompt. The results are reported in Table \ref{tab:pred_lag}. We see that the deep models show an improvement over logistic regression. The transformer model performs the best and particularly shows an improvement in modeling long sequences (N = 15, 25). We next present the results using the transformer model with pre-training. The EMA transformer model is pre-trained as described in Section \ref{sec:pt}. The model is then fine-tuned for the non-response prediction task and the results are in Table \ref{tab:pt_bert}.

\begin{table}[h]
        \centering
        \begin{tabular}{|c|c|c|c|c|}
            \hline
             Model & N = 5 & N = 10 & N = 15 & N = 25\\
             \hline
             Logistic regression & $0.70 \pm 0.03$ & $0.66 \pm 0.02$ & $0.65 \pm 0.02$ & $0.58 \pm 0.02$\\
             \hline 
             Vanilla LSTM & $0.74 \pm 0.02$ & $0.74 \pm 0.01$ & $0.73 \pm 0.02$ & $0.72 \pm 0.01$\\
             \hline
             Attention LSTM & $0.73 \pm 0.02$ & $0.73 \pm 0.02$ & $0.72 \pm 0.02$ & $0.71 \pm 0.01$\\
             \hline
             EMA transformer & ${0.75 \pm 0.02}$ & ${0.76 \pm 0.01}$ & ${0.76 \pm 0.01}$ & $0.75 \pm 0.01$ \\
             \hline
        \end{tabular}
        \caption{Average 5 fold cross validation AUC for predicting non-response to next EMA using a sequence of N EMAs. The transformer model here is directly trained for the task of non-response prediction without any pre-training. \textit{Note that the transformer model here is learned directly on the non-response prediction task without any pre-training.} }
        \label{tab:pred_lag}
\end{table}

\begin{table}[h]
        \centering
        \begin{tabular}{|c|c|c|c|c|}
            \hline
             Model & N = 5 & N = 10 & N = 15 & N = 25\\
             \hline
             EMA transformer & {${0.75 \pm 0.01}$} & ${0.77 \pm 0.01}$ & ${0.77 \pm 0.01}$ & $0.75 \pm 0.02$\\
             \hline
        \end{tabular}
        \caption{EMA transformer with self-supervised pre-training. The pre-training task is to impute one item (`Happy') in the masked positions in the sequence. This pre-training task performed the best compared to the others.}
        \label{tab:pt_bert}
\end{table}

\subsection{Results with different pre-training tasks}

The results presented in Table \ref{tab:pt_bert} are with the pre-training task of imputing one emotion item. In Section \ref{sec:pt} we describe pre-training tasks ranging from imputing one emotion item at a time to imputing all emotion items. For example, in a sequence of 10 EMAs we mask the response to the emotion `Happy' at positions 1 and 5. The model is then trained to impute the value of `Happy' at these positions. We compare different pre-training tasks and the final fine-tuning performance to the non-response prediction task in Table \ref{tab:pt_task}. We see that the performance does not vary much with the pre-training task. 

\begin{table}[h]
        \centering
        \begin{tabular}{|c|c|}
            \hline
             Pre-training task & N = 15 \\
             \hline
             One item & {${0.76 \pm 0.01}$}\\
             \hline
             All items & {${0.75 \pm 0.02}$} \\
             \hline
             Imputing 5 items & {${0.76 \pm 0.02}$} \\
             \hline
        \end{tabular}
        \caption{Cross validation AUC for predicting non-response to next EMA prompt using a sequence of N EMAs. The transformer model is pretrained on different tasks here: 1. Imputing one emotion item at a time, 2. Imputing all the emotion items, 3. Imputing five emotion items.}
        \label{tab:pt_task}
\end{table}   




\subsection{Learned attention weights}

\begin{figure}
    \centering
    \includegraphics[scale=0.2]{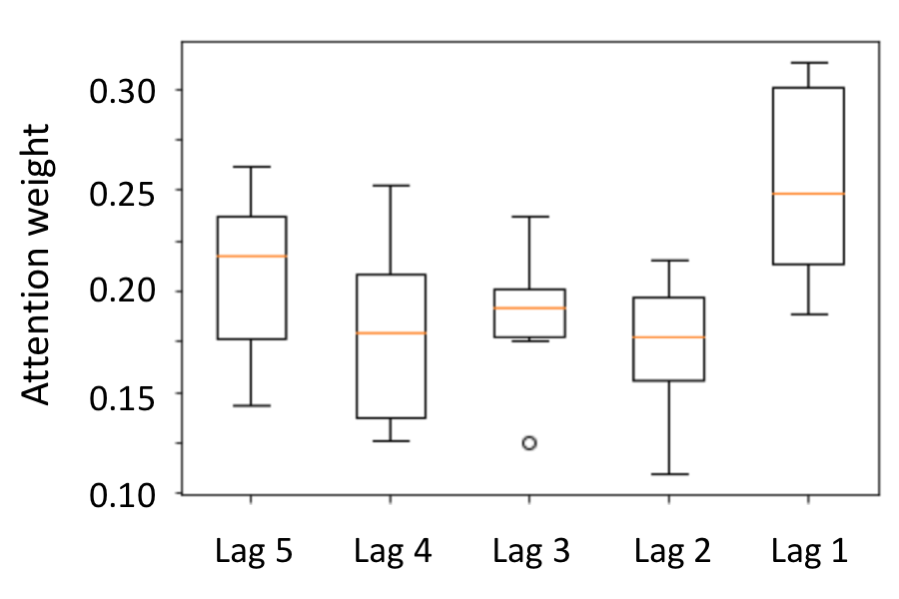}
  \caption{Attention weight across the different encoder layers in the transformer on the different EMA responses in a sequence of 5 EMAs in predicting non-response to the $6^{th}$ EMA. The x-axis contains the lag of the EMA with respect to the $6^{th}$ EMA. For example, \textit{Lag 1 is the most recent EMA}.}
  \label{fig:att_t}
\end{figure}

\begin{figure}
    \centering
    \includegraphics[scale=0.2]{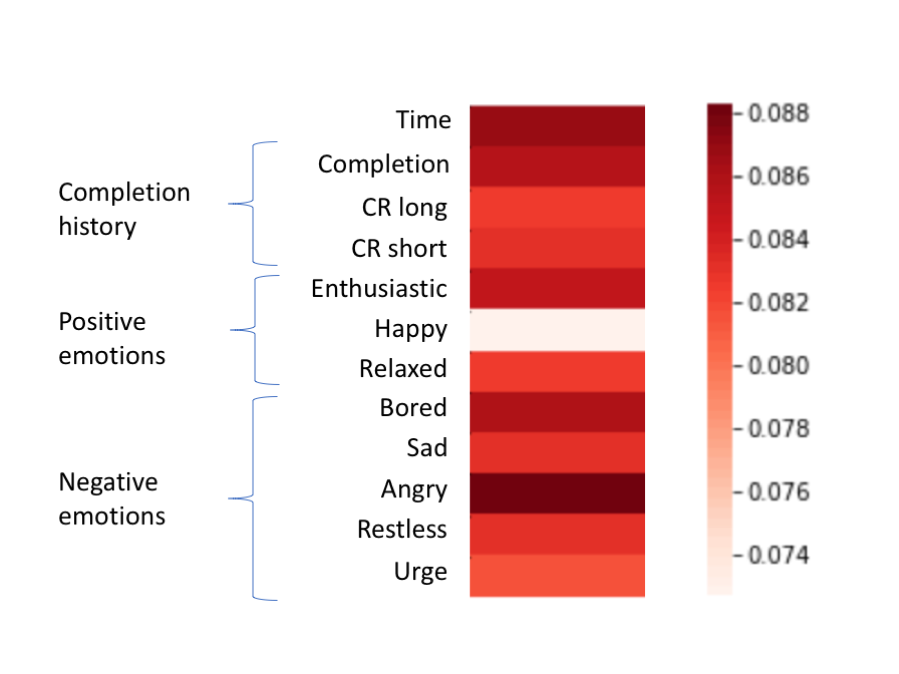}
    \caption{Attention weights on the different features in the input layer (first encoder layer). This provides an interpretation of the feature importance for the task of non-response prediction.}
  \label{fig:att_feat}
\end{figure}

The transformer encoder layers each perform multihead attention. The attention operation as described previously involves computing $A$ = softmax($\frac{QK^\top}{\sqrt{d_k}}$). The matrix $A$ is multiplied with the value matrix $V$. The columns of $A$ determine the weight or scaling factor for each row in $V$ which corresponds to the position in the EMA sequence. We compute the matrix $A$ from each encoder layer and plot the attention weight corresponding to each position in an EMA sequence of length 5. Figure \ref{fig:att_t} illustrates the weight on the different EMA positions in the sequence for predicting non-response to the next EMA. We also present the attention weights on the input features. We visualize the weight corresponding to each feature while computing the value matrix (with $W^V$). We present this analysis only for the first encoder layer since the input to subsequent layers is a linear combination of the input features and hence not interpretable at the feature level. This result is presented in Figure \ref{fig:att_feat}. We discuss the interpretation of these results in Section \ref{sec:discussion}.

\section{discussion}
\label{sec:discussion}

\subsection{Predicting non-response to EMA prompts}
We have defined a forecasting problem for prompt level EMA compliance. We evaluate the performance of classical statistical and machine learning models such as logistic regression and SVM (RBF kernel) for predicting next EMA compliance (compliance to $(N+1)^{th}$ EMA) using the current EMA ($N^{th}$ EMA) features. {Our feature vector summarizes the EMA history by using average completion and affect variance}. We perform an experiment to remove all summary features computed from the EMA items. In this setting, we {use} only the information contained in one EMA response to predict next EMA compliance. {This reduces performance substantially}, as shown in Table \ref{tab:baseline_removefeat_nextema} column `Raw features only'. These findings suggest that the {EMA history summary is} important for predicting next EMA compliance. 

{Next}, we explicitly incorporate the history of EMAs into the prediction model by using a fixed sequence of EMAs (sequence of $N$ EMAs) to predict next EMA ($(N+1)^{th}$ EMA) compliance. We develop a transformer based model for the task of forecasting using the sequence of EMA features. We see in table \ref{tab:pred_lag}
that using a sequence of EMAs {improves} performance (in LR and SVM when compared to using N = 1). Among the models predicting compliance from the sequence of EMAs, the EMA transformer performs the best.  

We hypothesize that the modest improvement in performance with deep learning models over classical models is due to the small size of our dataset. This prompts a couple of future directions for our work: 1) exploring domain adaptation methods to bridge the gap between several mHealth datasets and increasing the training data, 2) exploring transfer learning from other data domains (EHR,NLP) to improve the performance of deep models.

\subsection{Temporal/positional encoding for EMA input to transformers}

Since transformer models do not contain recurrence, the notion of position of words in a sentence is lost. The positional information is injected into the input sequence through positional encoding. In the canonical transformer works such as \cite{vaswani2017attention} and \cite{devlin2018bert}, positional encoding is performed by adding a function of the positional information to the input embedding. 

$$\text{Input to multihead attention} = \text{Input embedding} + PE(pos) $$ where $PE(pos)$ are sine and cosine functions of the position. 

In the case of EMAs, we have a more continuous notion of time when each EMA is triggered. Encoding the continuous time values into the input would provide a higher resolution of information than just the positions of each EMA. In the standard NLP literature, positional information is embedded by \textit{adding} sine and cosine functions of the position to the input. We believe that the \textit{additive} positional encoding strategy (used in NLP) is sub-optimal for EMA data where we have much smaller datasets. The idea here being that adding temporal/positional information changes the input values and the model has to learn to differentiate the effect of position/time and the actual input variation. In the case of NLP where datasets are larger, we hypothesize that the model can learn this distinction better. 

We encode temporal/positional information by concatenating it with the input embeddings. 
$$\text{Input to multihead attention} = 
           \begin{bmatrix}
           \text{Input embedding} \\
           PE(t)
           \end{bmatrix}$$

\begin{figure}
    \centering
    \includegraphics[scale=0.5]{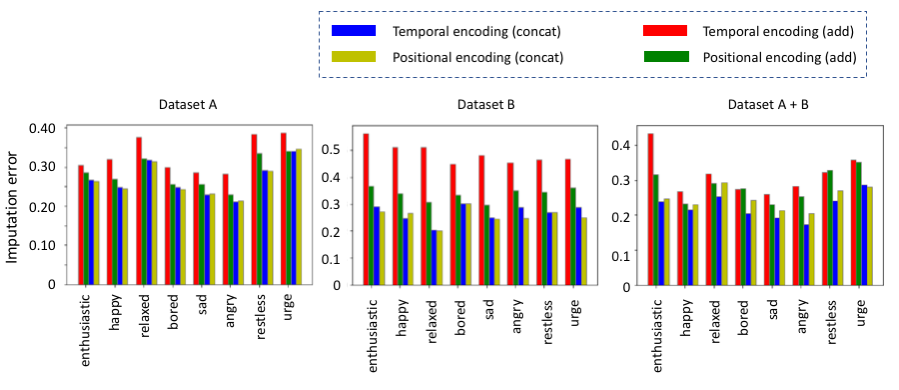}
    \caption{Pre-training task test performance (imputation error) when pre-trained with varying amount of data. We compare different temporal/positional encoding methods. We see that overall, concatenation performs better (lower error) than addition. 
    }
    \label{fig:pe_care_pns}
\end{figure}

The results for predicting non-response with different temporal/positional encoding strategies are shown in Table \ref{tab:pos_encoding}. This table also contains results when the size of the dataset used for pre-training the model varies. In the first two cases (smaller dataset), we see that concatenation performs better than additive encoding. In the third case (using slightly more data for pre-training), we see that the performance with additive positional/temporal encoding improves. We visualize the test results (imputation error) for the pre-training task of masked imputation in Figure \ref{fig:pe_care_pns}. The imputation error is the test error for imputing each emotion item. The results with four different positional encoding strategies are shown: Positional encoding (add), Temporal encoding (add), Positional encoding (concat), Temporal encoding (concat). We see that concatenation is always better than addition for the purpose of encoding temporal information.

\begin{table}[h]
        \centering
        \begin{tabular}{|c|c|c|c|c|}
            \hline
             Pre-training dataset & Positional (concat) & Temporal (concat) & Temporal (additive) & Positional (additive)\\
             \hline
             Dataset A only & {${0.76 \pm 0.01}$} & ${0.77 \pm 0.01}$ & ${0.70 \pm 0.01}$ & $0.73 \pm 0.02$ \\
             \hline
             Dataset B only & {${0.76 \pm 0.01}$} & ${0.77 \pm 0.01}$ & ${0.73 \pm 0.01}$ & $0.75 \pm 0.01$ \\
             \hline
             Dataset A + B & {${0.76 \pm 0.01}$} & ${0.76 \pm 0.01}$ & ${0.75 \pm 0.01}$ & $0.75 \pm 0.01$ \\
             \hline
        \end{tabular}
        \caption{Positional encoding strategies. All results reported are for N = 15}
        \label{tab:pos_encoding}
\end{table}

\subsection{Self-supervised pre-training of EMA transformer}

Pre-training transformer models in a self-suppervised manner has been shown to be beneficial in other domains such as NLP. The BERT model is pre-trained on a large corpus of text to learn structure in sentences. This model is then fine-tuned for new tasks where labeled datasets are of smaller sizes. Such a capability would be beneficial in fields such as mHealth where labeled datasets are difficult to collect. We evaluate the utility of pre-training and find a small improvement in the non-response prediction performance. Table \ref{tab:pt_bert} reports the AUC scores with a pre-trained transformer model. We see an improvement over the results in Table \ref{tab:pred_lag}. However, this isn't a statistically significant improvement. We hypothesize that this is due to the small size of our datasets. The question of the utility pre-training with EMA sequences as in the NLP setting remains a question for future work. However, the preliminary results that we obtain show that pre-training might have the potential to result in significant improvements with larger datasets.

\subsection{Interpreting the learned attention weights}

Figure \ref{fig:att_t} presents a visualization of the attention weights learned by the different transformer encoder layers on the EMA positions within a sequence. This corresponds to the importance of each EMA position within a sequence of $N$ EMAs in predicting non-response to the $(N+1)^{th}$ EMA. We see that Lag 1 has the highest weight across all of the encoder layers. Lag 1 corresponds to the most recent EMA, to the EMA for which we are predicting non-response. This is intuitive, as the most recent EMA is likely to be most closely-related to the paricipant's current mental state. We see an interesting pattern in the attention weights corresponding to the other lags. The mean of the lag 2, 3, and 4 weights are similar, but the weight corresponding to lag 5 is higher. One possible explanation for this trend is that since there are 4 EMAs on average per day, an EMA at lag 5 would correspond to the same time window as the current prediction task, but on the previous day. This may be capturing aspects of the participant's daily routine that are relevant to their response or non-response. More explicit methods for learning features derived from daily routines, diurnal rhythms, and related patterns is an interesting topic for future work.

In Figure \ref{fig:att_feat} we visualize the attention weights on the different features in the first encoder layer. This corresponds to the relative importance of different types of features in predicting non-response to the next EMA prompt. We see that time is an important feature in predicting non-response. This is also aligned with our finding that the way we encode time in the input affects the performance of the transformer model. We also find that the emotion features \emph{Enthusiastic}, \emph{Angry}, and \emph{Bored} have higher attention weights in comparison to other emotion items. We believe this is reasonable, as these are examples of strongly positive and negative emotions, which could influence response behavior, as well as the feeling of boredom which may correlate with a lack of engagement. The completion feature captures the detailed pattern of completion to each EMA in the sequence, while CR long and short are the average completion rate features that capture a summary of the pattern of completion. We see in Figure \ref{fig:att_feat} that the detailed pattern of completion is more useful in predicting non-response when compared to the long and short summaries of completion. This suggests that the model gets significant benefit from modeling the more detailed patterns in the response history. Collectively, these visualizations provide qualitative evidence that the transformer model is capable of learning meaningful structure from the sequence of EMA responses. The ability to identify and visualize the feature interactions learned by the transformer can be a potentially useful capability for domain scientists who are interested in designing related interventions.


\section{conclusion}

In this paper, we present a transformer architecture for modeling EMA sequences to predict non-response to future EMA prompts. Existing work on analyzing and predicting non-response have used classical machine learning models for this task. We are the first to explore the use of transformers for modeling sequences of  irregularly sampled EMA responses and predicting non-response to future EMA prompts. We address three issues in this work: 1. Choice of the input representation for EMA sequences, 2. encoding the temporal information into the input, 3. analyzing the utility of self-supervised pre-training on EMA data for improving the non-response prediction task. We find that the transformer model achieves a classification AUC of 0.77 and outperforms both classical ML and LSTM based DL models. We find that the design choice for positional/temporal encoding affects the performance of the model. We find that concatenating the temporal information leads to better performance when compared to the standard practice of adding the positional embedding. We design a self-supervised pre-training task on EMA sequences and find that it leads to a modest improvement that is not statistically significant. We present visualizations of the learned attention weights that illustrate the ability of the transformer to learn meaningful representations. We will make the predictive model trained from a corpus of 40K EMA samples freely available to the research community. An important future step will be using these prompt level compliance forecasts to inform the timing of compliance interventions.

\begin{acks}

\end{acks}

\bibliographystyle{ACM-Reference-Format}
\bibliography{references}
\appendix
\section{Appendix}
\label{sec:appendix}

\subsection{Supplementary figures}
\label{sec:appendix_fig}
\begin{figure}
    \centering
    \includegraphics[scale=0.2]{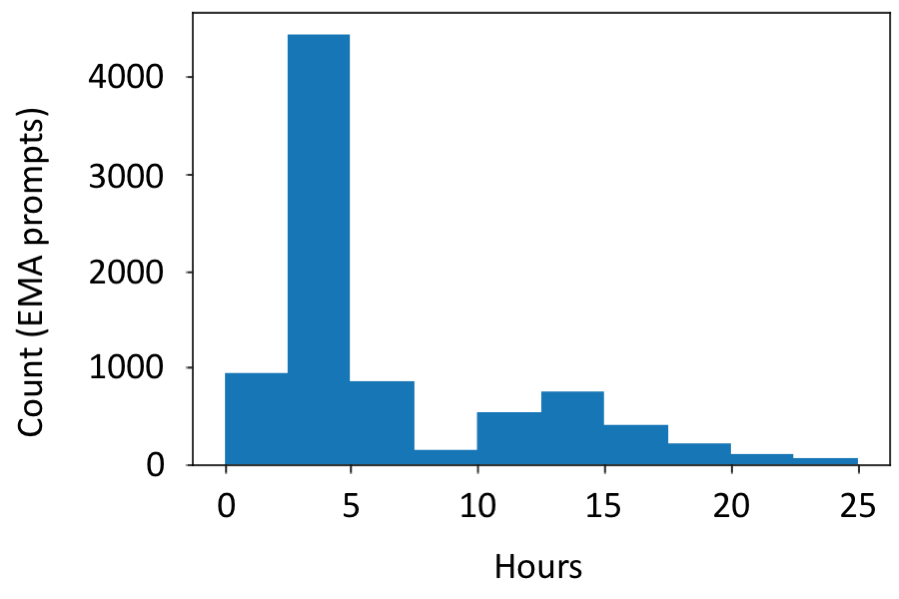}
    \caption{Histogram of the time between two consecutive EMA prompts. We see that the time difference is around 4 hours in most cases. The exceptions are when the app determines that the participant is unavailable and triggers an EMA at a different time.}
    \label{fig:time_diff_prompt}
\end{figure}

\begin{figure}
    \centering
    \includegraphics[scale=0.4]{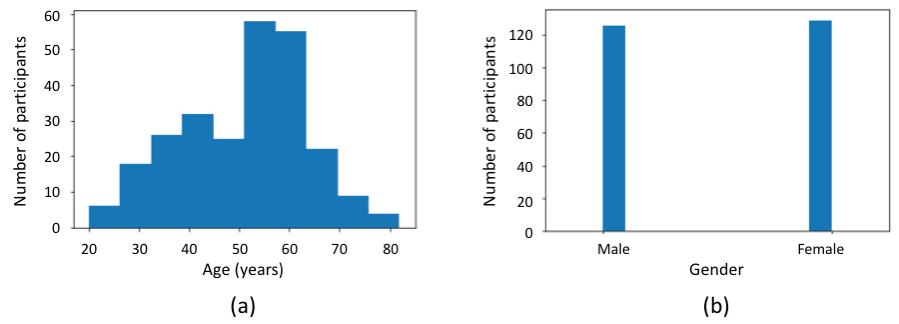}
    \caption{Demographics distribution in our dataset. (a) Age (b) Gender}
    \label{fig:age_gender}
\end{figure}


\end{document}